\documentclass[11pt]{article}
\usepackage[a4paper,margin=2.2cm]{geometry}
\usepackage[T1]{fontenc}
\usepackage[utf8]{inputenc}
\usepackage{lmodern}
\usepackage{booktabs}
\usepackage{array}
\usepackage{microtype}
\usepackage[hidelinks]{hyperref}
\hypersetup{pdftitle={ufakzeka-karar: An Open Turkish Typed-Decision Model with Order-Invariant Option Scoring},pdfauthor={Sait Furkan Teke}}
\usepackage{xurl}
\usepackage{enumitem}
\usepackage[font=small,labelfont=bf]{caption}
\setlist{nosep}
\widowpenalty=10000\clubpenalty=10000\title{ufakzeka-karar: An Open Turkish Typed-Decision Model\\ with Order-Invariant Option Scoring}
\author{Sait Furkan Teke\\ ufak AI\\ \texttt{furkan@ufakai.com}}
\date{October 2026}
\begin{document}
\maketitle

\begin{abstract}
ufakzeka-karar is an open Turkish decision model with 182,494,466 parameters. Given a Turkish text and questions of a fixed answer type (a choice, a level on an ordered scale, or yes or no), it returns a temperature-scaled probability for every option and an expected error that serves as a ``not sure'' signal, without generating text and in one CPU forward pass for up to ten options. Built on the lab's ufakzeka-1-base, its head scores each option blind to the others at shared positions, so the answer does not depend on option order. A sequential head trained with shuffled options was about as accurate but changed 2.3 to 2.8 percent of its answers when only the option order changed; REINFORCE lost 10.2 points (0.102) of macro F1 to cross-entropy. On the open set of HakemBench v1.0 (4,275 questions, 7 tracks) the released model ranks 7th of 16 rows with a composite of 0.660 (95\% interval 0.642 to 0.677). Temperature scaling lowers calibration error (smooth ECE) on the development set but raises it on held-out support questions, from 0.027 to 0.045 for the first scored run, which never trained on them; the released model later trained on them, so its 0.036 to 0.064 is not an unseen-question test. The released model is the last of three runs scored on HakemBench, and its numbers are not blind. The second run's new training data was aimed at the first run's errors on the full test set in guardrails, moderation and customer support, and the released run was trained after the second run's guardrail results on the full test set were read, under a protocol fixed in writing before any of its data, code or runs. All its numbers come after these readings; its guardrail, moderation and customer support numbers carry the flag ``shaped by reading the test results''. With every model scored on the other four tracks only, its composite is 0.678, 6th of 16. Weights and code are under Apache-2.0.
\end{abstract}

\section{Introduction}
Language models are also used for decisions with a fixed set of answers, such as routing a support ticket, flagging a phishing message, grading an answer or stopping a prompt injection. A typed-decision interface, defined by the hosted decision API Jev \cite{typesafe} and followed by open models built for English or for many languages at once \cite{kev,laya,decider,openjev}, fixes the answer type in advance and returns a probability for every allowed answer. The probabilities are the product: a caller acts on the answers above a threshold and sends the rest to a person \cite{geifman}, which works only as well as the probabilities are calibrated. A CPU model can run on the caller's own machine, so the text need not leave it.

ufakzeka-karar is a small model of this kind for Turkish, built on the lab's first model \cite{ufakzeka1}. We make four contributions: (i) an open model that answers a typed question in a median of 102.5 ms on one CPU thread; (ii) the order-invariant construction of Set-LLM \cite{setllm} applied to a typed-decision head and measured against reinforcement learning and against a sequential head that reads the options one after another and was trained with shuffled options; (iii) calibration measured on support questions held out of Run 1's training, including where it fails; and (iv) a comparison on HakemBench, the Turkish typed-decision benchmark of a companion paper \cite{hakembench}, with its disclosures: the model ranks 7th of 16 rows, and its training data and final run were shaped by reading earlier runs' results on the full test set. Result numbers trace to result files committed in the project's repository; internal audit figures, such as the hand-checked samples, are reported as recorded. Each selection and acceptance rule was written down before its result was seen. Square brackets give 95 percent bootstrap intervals throughout.

We follow Jev's public contract \cite{typesafe}: up to 255 options for a choice question and 2 to 10 ordered levels for a score question. Kev \cite{kev} trains low-rank adapters and a pointer head on Qwen3.5 base models (its 4B and 9B models are on the board) and keeps each question isolated from the others after the shared text; Laya \cite{laya}, a non-autoregressive engine on ModernBERT-large and mmBERT-base encoders, reads a question's options together in one input; decider-2b \cite{decider} and open-jev \cite{openjev} are further open models. Answers of large language models (LLMs) change when options are reordered \cite{pezeshkpour}. On HakemBench's permutation probe, order sensitivity among the other measured models runs from 0.039 (Jev 1.13) to 0.590 (Laya) \cite{hakembench}. Set-LLM \cite{setllm} gives a pretrained decoder a mask and position encoding under which set elements neither see each other nor differ in position, proves invariance and reports no accuracy cost, while scoring options in isolation did not reliably raise accuracy in multiple-choice evaluation \cite{isolation}. Supervised fine-tuning gave better calibration than reinforcement learning with verifiable rewards on decision tasks \cite{rlvrcal}; we selected on validation loss after temperature scaling \cite{guo} and calibrated afterwards, as Berta et al.\ recommend \cite{refine}.

\section{Model}\label{sec:model}
\paragraph{Backbone.} ufakzeka-1-base \cite{ufakzeka1} has 24 layers and hidden size 768, uses a Turkish tokeniser of 40,960 entries, was pretrained on 13.5B tokens and is read causally. Of the released file's 182,494,466 parameters, 182,492,928 are the backbone's, 769 the scoring layer's and 769 an abstain layer's that the release does not use; without the 31,457,280-parameter embedding matrix it has 151,037,186, ufakzeka-1's ``151M'' convention.

\paragraph{Input and scoring.} A question is one sequence: a prefix holding the text, \emph{Soru:} and the question, then each option as its own segment (\emph{Seçenek: $\langle$name$\rangle$. $\langle$description$\rangle$}; a level as \emph{Düzey $k$: \dots}; yes and no as \emph{Cevap: evet} and \emph{Cevap: hayır}). The prefix holds at most 448 tokens and an option at most 48. The prefix attends only to itself and each option causally to the prefix and itself, never to another option, and every option starts at the position after the prefix. Each option's hidden states are mean-pooled and scored by one linear layer, and a softmax runs over the options. Options are laid out sorted by their encoded text and scores are returned in the caller's order, so even summation order cannot change a bit; the repository's permutation test on the real weights demands bit-for-bit equality. More than ten options are split in sorted order into passes of ten under one softmax; since an option's logit does not depend on its pass-mates, the split does not change the answer, and reordering changes the probabilities only by floating-point noise.

\paragraph{Objective.} Cross-entropy against each row's target distribution, soft where labellers disagreed, plus on score questions the ranked probability score \cite{epstein}, so mass two levels off costs more than one; both are strictly proper. Rows are weighted to each task's natural answer mix; no reinforcement learning.

\section{Data}\label{sec:data}
Training data comes only from sources under Apache-2.0, MIT, CC0, CC BY 2.0 or CC BY 4.0, or written for the project; a licence manifest lists every file and the build fails on a missing or disallowed entry. Converted labelled sets keep their own labels, 3,000 rows each in the released mix: MASSIVE 1.1 in Turkish \cite{massive} (CC BY 4.0), the Turkish OffensEval 2020 corpus \cite{offenseval} (annotations CC BY 2.0), MiDe22 \cite{mide22} (MIT), Constitutional Court individual-application decisions \cite{aym} (CC BY 4.0), FACTurk \cite{facturk} (MIT) and WebFAQ \cite{webfaq} (collection CC BY 4.0, labelled from page structure). Three prompt-injection sets \cite{pi1k,pihn,piberat} (CC BY 4.0 and Apache-2.0; 1,791 rows) are labelled by the class each text was written or translated for. The released run added a conversational prompt-injection set (CC BY 4.0; 530 rows) whose texts are synthetic, model-written and curated by its author \cite{piconv}. For some sources the licence covers the collection, while the rights to the individual texts stay with their authors or the sites that published them.

For customer support, question templates in eight families (topic, urgency, handover to a person and others), written by various LLMs, were applied to 6,000 question-and-answer pairs from Turkish FAQ and community pages in clips/mqa \cite{mfaq} (collection CC0 1.0), 5,489 of them after a later filter, giving 12,957 training rows. The target is the mean of the probability vectors of two judge models from two families, read with the options rotated; they agreed on the top answer on 81 percent of training rows, and on the 255 development questions the panel agreed with the human answers on 75.3 percent [69.8, 80.4]. The human answers come from one person, the author; one annotator makes mistakes too, so figures that rest on them are indicative. The four support questions HakemBench asks of every support item (2,311 evaluation rows) were held out of Run 1's training to test generalisation to unseen questions; Run 1 saw their three families only in other question types. Generated tracks (spam and phishing messages, student answers, court-routing scenes, posts with and without checkable claims; 10,250 rows) were written by two models and labelled by two LLM judges other than the writer (1,531 check-worthiness rows ask about FACTurk claims or clips/mqa questions instead), since we found no licensed human-labelled Turkish set for them. Every file was checked against 70 evaluation sets (328,398 items) and every HakemBench v1.0 item, probe and development text, by shared 8-grams and MinHash near-duplicates \cite{broder} in both directions, and flagged rows were removed. The mix takes at most 3,000 rows a task. Run 1 trained on 42,998 rows and Run 2 on 55,870; Run 3, the released run, trained on 55,679, 12,151 of them from Run 2's additions.

\section{Training and selection}\label{sec:train}
\paragraph{Recipe.} AdamW \cite{adamw}, learning rate $10^{-4}$ for the backbone and $10^{-3}$ for the head, weight decay 0.01, 6 percent linear warm-up then linear decay, gradient clipping at 1.0, 32 questions a step, 2 epochs. The released run took 3,480 steps on one NVIDIA H200, with 427.8 seconds of training, 692.9 seconds in all and 13.5 GB peak memory. Every recipe ran with three random seeds; seeds and checkpoints were chosen on a support development set of 255 support questions with human answers, kept apart from the test set. Two earlier development rounds served the ablations (Section~\ref{sec:ablations}); Table~\ref{tab:rounds} compares the three runs scored on HakemBench, Runs 1 to 3. Run 1 was trained on the data as cleaned when the benchmark was frozen, never on the held-out support questions, and scored before anything was published. Its results on the full test set were read; on guardrails it caught 205 of 217 attacks but passed only 70 of 201 harmless look-alikes (attack-like messages).

\paragraph{Run 2} added data aimed at Run 1's errors on the full test set in guardrails, moderation and customer support. It asked the four support questions on 2,697 texts from the training pool and 300 from the validation pool, labelled by two AI models of one family, added 800 harmless guardrail messages in twenty product contexts (721 for training, 79 for validation), and added 1,499 comments (899 offensive, 600 clean) labelled blind by a model other than their writer. Run 2 passed its acceptance check (0.991 macro F1 on 224 validation rows of the earlier prompt-injection sets; Run 1 0.982), yet on HakemBench it caught only 106 of 217 attacks and passed 195 of 201 look-alikes. Every training attack was a template or a short translated prompt, and the only guardrail texts in a real user's register were Run 2's harmless ones: a likely register shortcut \cite{shortcut}, invisible to a check drawn from the same sets as the training attacks. Run 2's data has three costs. The new support questions are the four held-out questions, and the held-out file shares 887 of its 937 texts with the validation pool, so support is no longer a generalisation test. Their labels come from the AI model family that settled the benchmark's support gold under the same rubric, so a support score there is partly agreement with the labeller. And the new comments resemble the benchmark's model-written moderation items, all in the open set (a style check: a character n-gram classifier trained on them reached 0.926 balanced accuracy on those items but 0.401 on web-sourced moderation texts, against 0.818 and 0.619 for one trained on the older rows), so the moderation results of Runs 2 and 3 may read too high.

\paragraph{Run 3 and the flag.} Run 2's guardrail results on the full test set were read, and they showed the shortcut. Because training again after seeing test results weakens what the test measures, the protocol of Run 3 was fixed in writing before any of its data, code or runs; the selection rule and its result ship with the code. It added the train split of the conversational prompt-injection set \cite{piconv} and made its validation and test splits (50 attacks, 170 harmless messages) a guardrail development set. Two variants ran with three random seeds each, one with half of Run 2's 721 harmless look-alike rows and one without them. A run qualified if it caught at least 80 percent of the development attacks (yes probability at least 0.5); the qualifying run with the highest macro F1 on the support development set would be scored once, with no tuning after it. Four of six runs qualified and the rule chose a run without the look-alike rows, which caught 42 of 50 development attacks (recall 0.84 [0.725, 0.933]), passed 161 of 170 harmless messages and reached a development macro F1 of 0.688. Every number of Run 3 therefore comes after two readings of the test results, Run 1's in all three targeted tracks and Run 2's in guardrails, so its guardrail, moderation and customer support numbers carry the flag ``shaped by reading the test results''.

\section{Calibration}\label{sec:calib}
\paragraph{Method.} Validation rows were split 80/20 by a hash of the text into fit and selection parts, development items removed. Temperature scaling \cite{guo} in four forms (none, global, per question type, per type and option-count bucket) was fitted by cross-entropy on the fit part, and the form with the lowest cross-entropy on the selection part was chosen, preferring a simpler form within 0.001: for the released run one global $T = 1.2614$. The ``not sure'' value is the expected error given the raw maximum probability, an isotonic map \cite{isotonic} fitted on the selection part, the error being one minus the target's mass on the model's answer. No other confidence score beat maximum probability \cite{msp} in a check on an earlier checkpoint (area under the generalised risk-coverage curve, AUGRC, 0.1635 against 0.1642 for DOCTOR \cite{doctor}), in line with Traub et al.~\cite{augrc}. We report the Brier score \cite{brier} and the smooth expected calibration error (smooth ECE) \cite{smoothece}, a kernel-smoothed gap between stated confidence and observed accuracy; lower is better.

\begin{table}[t]
\centering\footnotesize\setlength{\tabcolsep}{2.5pt}
\begin{tabular}{lrlcccrcccc}
\toprule
 & & & attacks & look-alikes & human & & \multicolumn{2}{c}{held-out rows} & \multicolumn{2}{c}{support dev.\ set}\\
\cmidrule(lr){8-9}\cmidrule(lr){10-11}
run & composite & interval & caught & passed & check & $T$ & raw & with $T$ & raw & with $T$\\
\midrule
Run 1 & 0.556 & [0.537, 0.574] & 205 & 70 & 0.405 & per type & 0.027 & 0.045 & 0.051 & 0.034\\
Run 2 & 0.645 & [0.628, 0.663] & 106 & 195 & 0.524 & 1.2932 & 0.036 & 0.077 & 0.062 & 0.033\\
Run 3 (released) & 0.660 & [0.642, 0.677] & 194 & 128 & 0.476 & 1.2614 & 0.036 & 0.064 & 0.050 & 0.038\\
\bottomrule
\end{tabular}
\caption{The three training runs scored on HakemBench (Run 3 is released): composite on the open set (Section~\ref{sec:eval}) with its 95 percent bootstrap interval, guardrail attacks caught (of 217) and harmless look-alikes passed (of 201), accuracy on the human check (84 support questions answered by hand without seeing the labels), each run's own temperature $T$ (Run 1 fitted one per question type: 1.2341 for choice, 1.4003 for yes or no and 1.0507 for score questions), and smooth ECE (lower is better) without (raw) and with $T$ on the 2,311 held-out support rows, whose four questions Run 1 never trained on and Runs 2 and 3 did, and on the support development set (dev.\ set).}
\label{tab:rounds}
\end{table}

\paragraph{Results.} Temperature scaling makes calibration worse on a set of held-out support questions: for Run 1, which never trained on them, its own temperature raises smooth ECE from 0.027 to 0.045 (Table~\ref{tab:rounds}). The released model later trained on those four questions (Runs 2 and 3 add them), and its temperature was fitted on validation rows that ask those questions on texts from the same pool, so its own rise from 0.036 to 0.064 on that set (Brier 0.274 to 0.278) is not an unseen-question test; Run 2 moves the same way, and no interval was computed for these changes. On the support development set the temperature lowers the released model's calibration error from 0.050 to 0.038. A likely reason is that a temperature fitted on mildly overconfident validation data overcorrects on questions at a different confidence level. The selection rule was fixed before the held-out result was seen, so $T = 1.2614$ ships as chosen; $T = 1$ would have kept its held-out calibration error at 0.036. On the benchmark the model stays somewhat overconfident (the post-hoc temperature the benchmark fits for it is 1.1733).

\paragraph{The not-sure signal on HakemBench.} With confidence read as one minus the not-sure value, a threshold of 0.9 lets through 0.292 of the 4,275 open questions (1,248), and 0.062 of those answers are wrong. At 0.99 the error is higher, 0.085 on 213 answers: 148 of them are guardrail answers, 18 of those are wrong (0.122, Wilson interval \cite{wilson} 0.078 to 0.184), and no other track has an error there. The signal cannot be relied on to catch guardrail errors. On customer support almost no answer reaches 0.9 (6 of 1,340). No other interval was computed for these figures.

\section{Evaluation}\label{sec:eval}
The results below are those of the released model, Run 3. As Section~\ref{sec:train} describes, its numbers are not blind: every one comes after readings of the first two runs' results on the full test set, and its guardrail, moderation and customer support numbers carry the flag ``shaped by reading the test results''. Figures are on HakemBench's open set unless the text names another set.

\paragraph{HakemBench.} The open set of HakemBench v1.0 \cite{hakembench} holds 2,346 items and 4,275 questions in seven tracks. Most gold was settled by passes of AI models, and 260 questions carry human-decided gold. The composite, by which the public leaderboard (the board) ranks models, is the equally weighted geometric mean of three axes averaged over tracks: decision quality (macro F1; ``intelligence'' in the board files), calibration (Brier score relative to the uniform guess) and selective automation (AUGRC relative to a model wrong on every question). Intervals, unless named otherwise, are 95 percent percentile intervals from 2,000 bootstrap resamples over items, the same for every model; hosted models were queried on 26 and 27 September 2026 (UTC). Some named hosted models share a family with models that wrote or labelled parts of the benchmark data, and the companion paper also scores every model on the human-decided gold.

\begin{table}[t]
\centering\scriptsize\setlength{\tabcolsep}{3pt}
\begin{tabular}{rlrlrrrr}
\toprule
 & & & & decision & & selective & median\\
rank & model & composite & interval & quality & calibration & automation & ms\\
\midrule
1 & Gemini 3.8 Flash & 0.888 & [0.876, 0.898] & 0.901 & 0.806 & 0.964 & 1,294.2\\
2 & GPT-5.6 Sol & 0.842 & [0.827, 0.856] & 0.871 & 0.727 & 0.943 & 808.8\\
3 & GLM 5.3 & 0.827 & [0.813, 0.839] & 0.855 & 0.706 & 0.936 & 572.2\\
4 & Jev 1.13 & 0.825 & [0.811, 0.837] & 0.851 & 0.706 & 0.935 & 135.0\\
5 & Kev 4B & 0.688 & [0.673, 0.702] & 0.747 & 0.503 & 0.866 & 63.3\\
6 & Kev 9B & 0.666 & [0.647, 0.683] & 0.725 & 0.478 & 0.852 & 127.2\\
\textbf{7} & \textbf{ufakzeka-karar} & 0.660 & [0.642, 0.677] & 0.705 & 0.482 & 0.848 & 103.6\\
8 & Qwen3.5-4B & 0.653 & [0.629, 0.674] & 0.741 & 0.438 & 0.857 & 249.3\\
9 & DeepSeek V4 Pro & 0.633 & [0.606, 0.658] & 0.761 & 0.386 & 0.862 & 768.8\\
10 & decider-2b & 0.491 & [0.472, 0.510] & 0.587 & 0.273 & 0.737 & 48.3\\
11 & Laya multilingual (max\_len 1024) & 0.298 & [0.276, 0.317] & 0.425 & 0.113 & 0.547 & 116.4\\
12 & Laya multilingual (max\_len 2048) & 0.275 & [0.253, 0.297] & 0.418 & 0.093 & 0.533 & 12.2\\
13 & surface-cue baseline & 0.272 & [0.255, 0.291] & 0.329 & 0.111 & 0.553 & 0.0\\
14 & open-jev (DeBERTa-v3-large) & 0.234 & [0.213, 0.253] & 0.350 & 0.073 & 0.501 & 330.0\\
15 & Laya (max\_len 512) & 0.081 & [0.066, 0.100] & 0.233 & 0.006 & 0.366 & 369.4\\
16 & Laya (max\_len 2048) & 0.075 & [0.055, 0.095] & 0.221 & 0.005 & 0.365 & 15.1\\
\bottomrule
\end{tabular}
\caption{HakemBench v1.0 public board on the open set; axes as defined in the text, higher is better; interval: 95 percent bootstrap interval of the composite. Median ms: per question in the board run, not part of the composite, on setups that differ (a Mac CPU for ufakzeka-karar, open-jev and Laya at shipped length; an NVIDIA L4 GPU for Kev, Qwen3.5-4B and decider-2b; a GPU for Laya at full length; the network for hosted models; the surface-cue baseline runs no neural network, so its time is 0.0). Surface-cue baseline: a simple reference model we added for comparison that answers from cues such as length, digits or a question mark, not from meaning. max\_len: input length in tokens.}
\label{tab:board}
\end{table}

\paragraph{The board.} ufakzeka-karar ranks 7th (Table~\ref{tab:board}). Three hosted chat models (Gemini 3.8 Flash, GPT-5.6 Sol, GLM 5.3) and the decision API Jev 1.13 lead clearly (0.825 to 0.888), and Kev 4B and Kev 9B also score higher; the intervals of ranks 5 to 9 overlap its own, so that band is not settled. Its guardrail, moderation and customer support numbers are flagged. With every model scored on the other four tracks only, its composite is 0.678, 6th of 16 (no interval). On decision quality alone it is also behind Qwen3.5-4B (0.741) and DeepSeek V4 Pro (0.761), and on calibration (0.482) it is close to Kev 4B (0.503). DeepSeek V4 Pro's probabilities were read from its option labels' log-probabilities, so its calibration axis (0.386) reflects that route as well as the model.

\paragraph{By track.} Accuracy runs from 0.569 [0.534, 0.605] on customer support (1,340 questions; like guardrails and moderation, a flagged track) and 0.640 on fact-check triage to 0.952 on moderation; the public board file in the code repository gives every board model's macro F1 by track. The weakest places are the support choice and score questions (accuracy 0.490 and 0.481) and the triage priority scale (0.500, macro F1 0.268). The guardrail track has the highest calibration error, a smooth ECE of 0.158 against 0.025 to 0.084 on the other tracks; the model catches 194 of 217 attacks and passes 128 of 201 harmless look-alikes. On moderation it catches 150 of 157 offensive messages and passes 186 of 196 clean ones (see the style caveat of Section~\ref{sec:train}). The 181 court items and the 28 guardrail items taken from public prompt-injection sets are the test splits of datasets whose train splits are in ufakzeka-karar's training data (3,000 court rows, and 1,791 rows from three prompt-injection sets that include the guardrail items' two source sets; near-duplicates removed), so on those questions the model is supervised in-distribution while the other models answer zero-shot; its legal routing score (macro F1 0.754) and the guardrail counts should be read with that.

\paragraph{Support and the surface-cue baseline.} The surface-cue baseline answers from cues such as length, digits or a question mark, not from meaning. Over the whole open set it beats ufakzeka-karar on the support choice (0.516 against 0.490) and score questions (0.507 against 0.481), partly in-sample, as its cues were fitted on part of the open set. On the 84 support questions of the human check, answered by hand without seeing the labels, the model's accuracy is 0.476 [0.369, 0.583], below the baseline's 0.619 [0.512, 0.714], while the three leading hosted chat models and Jev 1.13 score 0.738 to 0.786 and DeepSeek V4 Pro 0.583.

\paragraph{Probes and speed.} The model's order sensitivity is 0.000 by construction. Half of its robustness value (0.941 [0.917, 0.966], Jev 1.13 0.943) is therefore 1.0 by construction; on the measured half, paraphrase agreement, it is 0.883 [0.833, 0.932] against Jev's 0.926 [0.889, 0.963], intervals overlapping. The English Brier gap (Turkish minus English Brier score; negative means worse in English) is $-0.303$ [$-0.377$, $-0.225$]; it was not trained for English. On an Apple M1 with one PyTorch thread and 13 sample questions, the model took a median of 102.5 ms a question (69.8 to 194.5 ms) with peak resident memory of 986,038,272 bytes in 32-bit floating point (fp32).

\paragraph{External Turkish test sets.} Each item was asked as a typed question with no examples and scored with each run's own calibration (Table~\ref{tab:external}). The MASSIVE, MiDe22 and OffensEval numbers are supervised, not zero-shot: their train splits are in the training data. MMLU-Pro-TR \cite{mmlupro,mmluprotr}, XCOPA \cite{xcopa} and X-FACT \cite{xfact} touch no training text, though all of X-FACT's Turkish claims come from Doğruluk Payı, one of FACTurk's sources. From Run 1 to Run 3 accuracy fell a little on the first four sets and rose on XCOPA and X-FACT; the intervals overlap on every set, and no paired test was run. On the knowledge exam MMLU-Pro-TR the released model is slightly below chance: 0.098 [0.092, 0.103] against 0.111.

\begin{table}[t]
\centering\footnotesize\setlength{\tabcolsep}{4pt}
\begin{tabular}{lrrrrlrrrr}
\toprule
 & & & \multicolumn{3}{c}{accuracy} & \multicolumn{2}{c}{Brier score} & \multicolumn{2}{c}{smooth ECE}\\
\cmidrule(lr){4-6}\cmidrule(lr){7-8}\cmidrule(lr){9-10}
set & $n$ & chance & Run 1 & Run 3 & Run 3 interval & Run 1 & Run 3 & Run 1 & Run 3\\
\midrule
MASSIVE 1.1 tr-TR & 2,937 & 0.017 & 0.756 & 0.745 & [0.728, 0.761] & 0.355 & 0.359 & 0.029 & 0.025\\
MiDe22 & 1,012 & 0.333 & 0.766 & 0.741 & [0.714, 0.767] & 0.334 & 0.350 & 0.035 & 0.030\\
OffensEval-TR 2020 A & 3,528 & 0.500 & 0.866 & 0.844 & [0.832, 0.857] & 0.204 & 0.225 & 0.021 & 0.020\\
MMLU-Pro-TR & 11,838 & 0.111 & 0.107 & 0.098 & [0.092, 0.103] & 0.936 & 0.941 & 0.126 & 0.135\\
XCOPA, Turkish & 500 & 0.500 & 0.562 & 0.584 & [0.542, 0.626] & 0.602 & 0.567 & 0.196 & 0.171\\
X-FACT, Turkish & 169 & 0.250 & 0.320 & 0.343 & [0.278, 0.414] & 0.826 & 0.749 & 0.181 & 0.072\\
\bottomrule
\end{tabular}
\caption{External Turkish test sets, Run 1 and Run 3 (released), each with its own temperature; $n$: rows scored; chance: accuracy of a uniform guess; Run 3 interval: 95 percent bootstrap interval of Run 3's accuracy; the first three sets are supervised (their train splits are in the training data; MiDe22 ships as one file, so its 70/10/20 split was drawn by the project before training); X-FACT is asked as Cetvel's four verdicts; OffensEval-TR 2020 A is subtask A, offensive or not. Dropped: 37 repeated MASSIVE test utterances, 2 MiDe22 rows overlapping training texts and 4 malformed MMLU-Pro-TR rows.}
\label{tab:external}
\end{table}

\section{Ablations}\label{sec:ablations}
Every ablation's compared choices (arms), measures and decision rule were fixed in writing before it ran; each training arm used three random seeds and was judged by a paired bootstrap over items within each task and over seeds, and the int8 files, built from one checkpoint, were compared with fp32 on the same rows (Table~\ref{tab:ablations}); for the head and objective arms the primary measure was macro F1 on the held-out questions, which neither the second development round nor Run 1 trained on, with calibration measures as secondary outcomes under a Holm correction \cite{holm}.

\begin{table}[t]
\centering\footnotesize\setlength{\tabcolsep}{4pt}
\begin{tabular}{>{\raggedright\arraybackslash}p{5.2cm}>{\raggedright\arraybackslash}p{5.4cm}>{\raggedright\arraybackslash}p{4.6cm}}
\toprule
compared choice & primary measure, points [95\% interval] & other findings\\
\midrule
masked-language conversion, 1B tokens & $-1.68$ [$-3.89$, $+0.33$] on the three general sets & adoption required $+1.0$; mean of four Turkish sets 0.7578 against 0.7598 for the causal backbone\\
BERTurk \cite{berturk}, same head and data & $-0.35$ [$-1.89$, $+1.69$] on decision tasks & $+1.67$ [$+0.04$, $+3.21$] on the three general sets\\
MoganBert-TR \cite{moganbert}, same head and data & $-0.79$ [$-2.20$, $+1.32$] on decision tasks & $-3.37$ [$-6.41$, $-0.11$] on held-out questions\\
continued pretraining, 300M tokens & $-1.26$ [$-2.44$, $+0.96$] & backbone kept\\
down-weighting easy rows \cite{cartography} & $-2.2$ [$-4.8$, $+0.3$] & rejected\\
sequential head, shuffled options & Run 1 $-1.7$ [$-4.3$, $+0.9$]; second round $-4.0$ [$-7.0$, $-1.0$] & 2.3 to 2.8\% of answers change under reordering\\
REINFORCE \cite{williams}, leave-one-out \cite{rloo} & Run 1 $-10.2$ [$-13.5$, $-6.2$]; second round $-12.7$ [$-15.5$, $-9.5$] & every calibration measure worse\\
int8 files, points lost & dynamic 1.42 [0.19, 2.71]; per channel 1.61; down projections in fp32 0.67 & all lose more than the allowed 0.5; smooth ECE within 0.11 points\\
\bottomrule
\end{tabular}
\caption{Ablations: the compared choice minus the shipped one in macro F1 points (hundredths of the 0 to 1 scale), unless stated. The conversion row ran in a first head trial on the causal backbone (general sets: MASSIVE, OffensEval and MiDe22), the BERTurk, MoganBert-TR and continued-pretraining rows in the first development round, the easy-row and int8 (8-bit integer weights) rows in the second, and the sequential-head and REINFORCE rows in Run 1 and the second round, as marked.}
\label{tab:ablations}
\end{table}

The backbone had been fixed as the lab's own before the encoder comparisons, so those are reported, not acted on. The sequential head's four-point loss in the second development round did not reproduce in Run 1, trained on cleaned data, so we claim only the smaller result, that the order-invariant head is as accurate and never depends on option order. REINFORCE (eight samples a question, the target's mass on the sampled answer as reward) has an expected reward linear in the probabilities, so its optimum puts all mass on one answer; cross-entropy's minimiser is the target. No int8 file stayed within the required 0.5 points of fp32 on macro F1, even with the 24 feed-forward down-projection matrices in fp32; the loss came from near-tie answers, so the release is fp32.

\section{Limitations}\label{sec:limits}
Six models on the board score higher. Support routing is weak: on the 84 support questions of the human check, the model is below the surface-cue baseline. All human answers come from one person; there is no inter-annotator agreement. Calibration does not transfer everywhere, and the not-sure signal misses guardrail errors. Every number of the released model comes after readings of earlier runs' test results, and its guardrail, moderation and support numbers are flagged. It passes only 128 of 201 harmless look-alikes. The moderation result may be inflated by style likeness. World knowledge is weak; the model reads Turkish only, and the end of a text beyond 448 tokens is cut. Most training labels come from LLMs or rules, and no prompt-injection or relevance label was checked by a person. The board's intervals resample items, but items a model wrote in one generation request are correlated, so they are likely too narrow.

\section{Cost and release}\label{sec:cost}
The project cost \$236.90 in GPU time and hosted model API calls (\$181.55 on Modal, \$55.35 in API calls). The figure does not include the server, the AI model used during development or the base model ufakzeka-1, whose cost is reported with it \cite{ufakzeka1}. The figure covers the model and HakemBench together and was read from the providers' dashboards.

Released under Apache-2.0: the weights with the calibrator and a small inference file that needs no transformers library, the code and the public result files. The release is at these addresses.\\[2pt]
{\small\begin{tabular}{@{}ll@{}}
Weights and card & \url{https://huggingface.co/ufakai/ufakzeka-karar}\\
Code & \url{https://github.com/ufakai/ufakzeka-karar}\\
Benchmark & \url{https://huggingface.co/datasets/ufakai/HakemBench}\\
Demo & \url{https://karar.ufakzeka.com}\\
Write-up & \url{https://ufakai.com/research/ufakzeka-karar}
\end{tabular}}

\paragraph{Use of AI tools.} An AI model was used throughout this project to write and run the training, data and evaluation code, to review plans, items and texts, and to draft this report. AI models also wrote and labelled training data (Sections~\ref{sec:data} and \ref{sec:train}), and AI passes settled most of HakemBench's gold and reviewed its probes \cite{hakembench}. The human answers on the support development set and in the human check are the author's own. The author directed the work, reviewed the results and text, and is responsible for all of it.

\end{document}